\documentclass[conference]{IEEEtran}
\IEEEoverridecommandlockouts
\usepackage[hyphens]{url}
\usepackage{cite,amsmath,amssymb,graphicx,xcolor,booktabs,array}
\usepackage{tikz}
\usetikzlibrary{arrows.meta,positioning,calc,fit,backgrounds,shapes.geometric}
\def\BibTeX{{\rm B\kern-.05em{\sc i\kern-.025em b}\kern-.08emT\kern-.1667em\lower.7ex\hbox{E}\kern-.125emX}}

\begin{document}

\title{Total Cost of Agency: Exact Attribution of Memory\\Injection Cost in Multi-Agent LLM Workflows}

\author{\IEEEauthorblockN{Vivek Kumar Singh}
\IEEEauthorblockA{\textit{Independent Researcher}\\
McKinney, TX, USA\\
vivekksingh.nov12@gmail.com\\
ORCID 0009-0002-9350-3207}
\and
\IEEEauthorblockN{Preeti Priyam}
\IEEEauthorblockA{\textit{Independent Researcher}\\
McKinney, TX, USA\\
preetipriyam12@gmail.com\\
ORCID 0009-0000-9423-741X}
\and
\IEEEauthorblockN{Gautam Bhowmick}
\IEEEauthorblockA{\textit{Independent Researcher}\\
Chicago, IL, USA\\
bhowmick.gautam@gmail.com\\
ORCID 0009-0001-9424-7826}}

\maketitle

\begin{abstract}
Every node in a multi-agent large language model (LLM) workflow retrieves context from memory and injects it into its prompt, where those injected tokens are billed as input tokens at the same per-token price as the system prompt and the user query. Production observability tools report total token cost but do not separate the tokens a node generates from the tokens it is handed, so this component of the bill is invisible to the teams paying it. We introduce the Total Cost of Agency (TCA), a decomposition of multi-agent workflow cost into base prompt, inference, memory injection, miss penalty and context-accumulation components, and an exact attribution method: a two-pass, non-billable token count that measures injected tokens directly rather than estimating them from word-count proxies. On a 200-task enterprise benchmark executed against real model application programming interfaces (APIs), memory injection accounts for 13.6 percent of the variable cost a compile-time optimizer can act on, approximately 12 percent of the full billed cost, and its share rises from a structural zero at workflow depth one to 27.6 percent at depth six. Injected tokens grow linearly with depth over the measured range ($R^2 = 0.9974$, depths two through six); a quadratic fit yields a negative leading coefficient, so the data do not exhibit convex growth at these depths. We show the component is controllable at fixed model tier: reducing the retrieval window capacity from 32 to 2 entries lowers injected tokens by 28.7 percent with an accuracy change within seed-level variation. We report in full that our graph-rewriting transforms are approximately cost-neutral in isolation, that two of the five decomposition terms are zero by construction in this harness, and that total workflow cost is dominated by model tier assignment, which we hold fixed and treat as orthogonal prior work. Prompt caching is not evaluated; all figures are for the uncached case.
\end{abstract}

\begin{IEEEkeywords}
memory injection, agent memory, prompt caching, token accounting, LLM agents, multi-agent systems, workflow cost, LangGraph
\end{IEEEkeywords}

\section{Introduction}
Running large language model (LLM) agents in production is expensive, and the expense is concentrated in places current tooling does not look. A frontier model is accurate but can cost twenty to twenty-five times what a small model costs per token, and real agent workflows rarely make a single call. They string several model invocations together in a dependency chain, and the cost of a multi-step workflow is dominated not only by which model each step uses, but by how much context each step must carry. This article describes a component of that cost that no production observability tool reports separately.

Consider a four-node enterprise billing reconciliation workflow. A planner decomposes a discrepancy analysis query into extraction, query generation, reconciliation and policy check subtasks, each issuing one or more model calls. Teams optimizing this workflow focus on model tier selection: they use a small model where possible and escalate to a larger one when needed. This is correct but incomplete. Every agent node also retrieves context from memory before inference and injects it into its prompt. Those injected tokens are billed as input tokens, at the same per-token price as the system prompt and the user query. In a workflow that carries the outputs of all prior nodes, the final node injects the accumulated context of the entire chain before generating any output of its own; Fig.~\ref{fig:flow} traces this accumulation and the accounting that follows from it.

\textbf{The measurement problem.} Production observability tools for LLM applications --- LangSmith~\cite{b23}, Arize Phoenix~\cite{b24} and Weights \& Biases Weave~\cite{b25} among them --- trace agent runs and report token counts and cost per call, but they report a call's input tokens as one quantity. They do not expose the split between tokens a node generated and tokens a node was handed. A team optimizing tier selection cannot reason about the second quantity, because in their instrumentation it does not exist as a separate number. The obstacle is not that the cost is hard to compute; it is that nothing in the standard stack computes it.

\textbf{Contributions.} This article makes three contributions. First, we formalize the Total Cost of Agency (TCA), a decomposition of multi-agent workflow cost into base prompt, inference, memory injection, miss penalty and context-accumulation components, and we state explicitly which components accrue billable cost under which deployment conditions. Second, we present an exact attribution method: a two-pass, non-billable token count that measures injected tokens directly using the provider's own tokenizer rather than estimating them from word-count or character-count proxies. The method costs nothing to run, adds roughly ten milliseconds per node against inference latencies of eighty to three thousand milliseconds, and yields the base-prompt cost as a by-product. Third, we report an empirical characterization across five workflow categories and five depths on a 200-task enterprise benchmark executed against real model application programming interfaces (APIs).

\textbf{What this article does not claim.} Total workflow cost in our experiments is dominated by model tier assignment, not by memory management. Routing work such as FrugalGPT~\cite{b1}, RouteLLM~\cite{b2} and MasRouter~\cite{b6} owns that result, and we hold tier assignment fixed and treat it as orthogonal prior work throughout. We also report, in Section~\ref{sec:negative}, that the graph-rewriting transforms we implemented are approximately cost-neutral in isolation. Our claim is confined to what we measured: that memory injection is a real, separately attributable and exactly measurable component of the bill, that it grows with workflow depth, and that it responds to a retrieval-capacity lever at fixed tier. Prompt caching, the most plausible mitigation, is not evaluated; Section~\ref{sec:limits} explains why the component is an unfavorable caching candidate here and states plainly that we did not measure it.

\section{Related Work}\label{sec:related}
\subsection{Cost-Aware Routing and Cascades}
The cascade idea originates with FrugalGPT~\cite{b1}, which demonstrated real savings on single-call natural language processing benchmarks by trying cheap models first and escalating when confidence is low. FrugalGPT models the cost of a request as a function of output tokens and per-token price; in its single-call setting there is no accumulating context to inject. RouteLLM~\cite{b2} and Hybrid LLM~\cite{b3} are query-level routers that dispatch an entire query to either a small or a large model based on predicted difficulty, without a notion of per-node memory operations. RouterBench~\cite{b4} standardizes the evaluation of such systems, and AutoMix~\cite{b5} escalates on self-verification failure. MasRouter~\cite{b6} learns routing policies for multi-agent systems and achieves substantial cost reductions on coding benchmarks, routing at the agent level with a learned policy. Cost-aware automation has also been explored at the pipeline level for machine learning workflows~\cite{b17}.

This body of work owns the model-selection lever, and in our experiments that lever moves total cost far more than anything in the memory subsystem does. What these approaches share is that cost is treated as a function of the model invoked, never of the context carried into it. We hold tier assignment fixed throughout and make no routing claim.

\subsection{Agent Memory, Context Compression and Prompt Caching}
MemGPT~\cite{b7} introduced a two-tier memory hierarchy for multi-session agents, paging information between a fast working context and a slower archival store to manage limited context windows. Its objective is memory accuracy and context freshness under a fixed window budget; it does not measure or minimize the token cost incurred when retrieved memory is injected into a prompt. Vector retrieval backends and the broader retrieval-augmented generation literature similarly optimize what is retrieved for relevance, not what the retrieval costs once billed as input tokens.

A complementary line of work reduces the cost of carried context by compressing it. Task-agnostic prompt compression such as LLMLingua-2~\cite{b18} distills prompts to a fraction of their original token count while preserving task fidelity, and key-value cache compression such as SnapKV~\cite{b19} reduces the memory and latency footprint of long contexts at serving time, building on paged key-value cache management at the serving layer~\cite{b22}. These techniques operate on a given prompt or cache; they are orthogonal to measuring what an injection costs, and compose with the accounting we describe.

Agent frameworks including LangGraph~\cite{b14} and AutoGen~\cite{b13} provide infrastructure for multi-agent workflows but leave model selection to external configuration and make the injection cost invisible. Reasoning patterns such as ReAct~\cite{b15} and Reflexion~\cite{b16} shape how much intermediate context accumulates, which is the quantity our benchmark prices. Execution-grounded and agent benchmarks~\cite{b8,b9,b10,b11,b12} evaluate task success but do not decompose cost.

Finally, providers discount cached input tokens substantially: a cache read is billed at $0.1\times$ the base input price, against $1.25\times$ for a five-minute write and $2\times$ for a one-hour write~\cite{b20}. Caching is the most plausible mitigation for the cost we measure. We do not evaluate it; Section~\ref{sec:limits} gives the structural reasons injected memory is an unfavorable caching candidate here, and the fact that we have not measured hit rates.

\section{The Total Cost of Agency}\label{sec:tca}
\subsection{Preliminaries}
We consider an agentic system that answers a query by executing a workflow $W = (V, E)$, a directed acyclic graph in which each node $v$ is a subtask paired with a specialist role and served by a language model drawn from a discrete set of tiers $T$. We write $p_{\mathrm{in}}(t)$ and $p_{\mathrm{out}}(t)$ for the input and output per-token prices of tier $t$. A directed edge $(u,v)$ indicates that $v$ consumes the output of $u$, so execution proceeds in topological order. The depth of a node is the length of the longest path from a source to that node, and the depth of a workflow is its maximum node depth. Before inference, each node retrieves context from memory under a memory strategy $m$, which determines how many prior outputs are injected into its prompt: a capacity-bounded strategy retains at most $K$ entries, while an unbounded strategy retains all prior history.

\subsection{The Decomposition}
We define the Total Cost of Agency of a workflow as the sum over nodes of five components:
\begin{equation}
\begin{split}
\mathrm{TCA}(W) = \sum_{v \in V} \big[ &C_{\mathrm{base}}(v) + C_{\mathrm{inf}}(v) + C_{\mathrm{inject}}(v)\\
 &+ C_{\mathrm{miss}}(v) + C_{\mathrm{accum}}(v) \big]
\end{split}
\end{equation}

$C_{\mathrm{base}}(v)$ is the cost of the tokens a node's prompt carries independently of memory --- its specialist system prompt, its tool definitions and the user query --- billed at $p_{\mathrm{in}}(\mathrm{tier})$. $C_{\mathrm{inf}}(v)$ is the inference cost, equal to the tokens the node generates times $p_{\mathrm{out}}(\mathrm{tier})$. $C_{\mathrm{inject}}(v)$ is the memory injection cost, equal to the number of memory tokens retrieved and injected into the node's prompt times $p_{\mathrm{in}}(\mathrm{tier})$. $C_{\mathrm{miss}}(v)$ is the penalty incurred when a warm-tier lookup misses and the system falls back to the durable store. $C_{\mathrm{accum}}(v)$ is the cost attributable to upstream context that a node inherits and carries forward.

We distinguish these terms by when they are non-zero, because a decomposition whose terms are unexercised is not informative. Three components are billed on every node in every configuration: $C_{\mathrm{base}}$, $C_{\mathrm{inf}}$ and $C_{\mathrm{inject}}$. Two are conditional. $C_{\mathrm{miss}}$ is non-zero only when the durable tier is a billed remote service; in a deployment whose durable tier is a local store, a warm-tier miss costs latency but no tokens, and the term evaluates to zero. $C_{\mathrm{accum}}$ is a conceptual rather than a separately measurable quantity: inherited context and retrieved context arrive in the prompt as the same tokens, and the attribution method of Section~\ref{sec:method} counts them together within $C_{\mathrm{inject}}$. We define both terms because a deployment with a billed durable tier will exercise the first, but we report throughout that neither accrues billable cost in the configuration we measure.

The component prior work omits entirely is $C_{\mathrm{inject}}$. The coupling that makes the accounting non-trivial is that $C_{\mathrm{inject}}$ scales with the input price of the tier a node executes on: the same injected context costs more at a more capable tier than at a cheaper one. A cost figure for injected context is therefore meaningful only alongside the tier that produced it, which is why Section~\ref{sec:results} reports injected volume in tokens as well as in currency.

\subsection{Expected Growth with Depth}
The following is analysis that motivates the measurements in Section~\ref{sec:results}; it is not offered as an empirical result. Consider a linear-chain workflow of depth $d$ in which each node injects the outputs of all prior nodes. Under an unbounded history strategy, node $i$ injects the outputs of nodes $1$ through $i-1$, contributing $(i-1)\tau$ tokens for an average per-node output length $\tau$. Summing over $i$ from $1$ to $d$ gives $\tau d(d-1)/2$, which is $O(d^2)$. Under a capacity-bounded strategy with window $K$, per-node injection is bounded by $K$ entries and total injection grows as $O(\min(K,d)\cdot d)$, which is linear in $d$ whenever $d$ exceeds $K$. For general directed acyclic graphs, $d$ is replaced by the maximum fan-in along the critical path.

Two statements about scope follow, and we make them explicit because the distinction governs how Section~\ref{sec:results} should be read. First, the quadratic expression describes the unbounded regime, which our experiments do not measure; we report the bounded regime only. Second, the capacity that bounds our baseline and the capacity we vary as an optimization lever in Section~\ref{sec:results} are the same mechanism at different settings, and we name them separately --- $K_{\mathrm{default}}$ for the harness configuration under which all conditions run, and $K_{\mathrm{opt}}$ for the setting varied in the capacity experiment --- so that no reduction attributable to lowering $K_{\mathrm{opt}}$ is mistaken for a property of the baseline.

\begin{figure*}[t]
\centering
\definecolor{memS}{HTML}{1668A8}\definecolor{memF}{HTML}{E2EDF6}
\definecolor{msrS}{HTML}{B0730A}\definecolor{msrF}{HTML}{FAEBD2}
\definecolor{msrH}{HTML}{F3D79B}
\definecolor{mdlS}{HTML}{7343B0}\definecolor{mdlF}{HTML}{EDE5F6}
\definecolor{accS}{HTML}{1F7D4A}\definecolor{accF}{HTML}{E1F0E7}
\definecolor{neuS}{HTML}{3A3A3A}\definecolor{neuF}{HTML}{F1F1F1}
\begin{tikzpicture}[
  font=\footnotesize, x=1cm, y=1cm,
  bx/.style={draw,rounded corners=1pt,minimum height=5.2mm,inner xsep=3pt,
             inner ysep=1pt,line width=0.4pt,align=center},
  nd/.style={bx,fill=black!5,minimum width=21mm},
  hi/.style={bx,fill=black!12},
  op/.style={draw,circle,inner sep=0.5pt,line width=0.4pt,minimum size=4.2mm},
  st/.style={draw=memS,fill=memF,cylinder,shape border rotate=90,aspect=0.13,minimum width=15mm,
             minimum height=6.5mm,inner sep=1pt,line width=0.4pt},
  ar/.style={-{Latex[length=1.6mm,width=1.3mm]},line width=0.45pt},
  da/.style={-{Latex[length=1.6mm,width=1.3mm]},line width=0.4pt,densely dashed,draw=black!55},
  tx/.style={inner sep=1pt},
  sm/.style={tx,font=\scriptsize},
  gy/.style={tx,font=\scriptsize,text=black!55},
  stage/.style={draw=black!45,densely dotted,rounded corners=2pt,line width=0.5pt},
  mem/.style={bx,draw=memS,fill=memF},
  msr/.style={bx,draw=msrS,fill=msrF},
  msrhi/.style={bx,draw=msrS,fill=msrH,line width=0.6pt},
  mdl/.style={bx,draw=mdlS,fill=mdlF},
  acc/.style={bx,draw=accS,fill=accF},
  amem/.style={ar,draw=memS}, amsr/.style={ar,draw=msrS,line width=0.55pt},
  amdl/.style={ar,draw=mdlS}, aacc/.style={-{Latex[length=1.6mm,width=1.3mm]},
    line width=0.45pt,densely dashed,draw=accS}
]

\node[tx,anchor=west,font=\footnotesize\bfseries] at (0.05,0.42) {1.~Compile};
\node[gy,anchor=west] at (1.55,0.42) {once, before execution};

\node[bx,minimum width=21mm] (q)  at (1.45,-0.30) {user query};
\node[bx,minimum width=21mm] (pl) at (1.45,-1.10) {planner};
\node[nd] (n1) at (1.45,-2.05) {extract\hspace{2mm}$d{=}1$};
\node[nd] (n2) at (1.45,-2.80) {SQL-gen\hspace{2mm}$d{=}2$};
\node[nd] (n3) at (1.45,-3.55) {reconcile\hspace{2mm}$d{=}3$};
\node[nd] (n4) at (1.45,-4.30) {policy\hspace{2mm}$d{=}4$};
\draw[ar] (q) -- (pl);  \draw[ar] (pl) -- (n1);
\foreach \a/\b in {n1/n2,n2/n3,n3/n4} \draw[ar] (\a) -- (\b);
\foreach \y/\n in {-2.80/1, -3.55/2, -4.30/3}
  \foreach \k in {1,...,\n}
    \node[draw,fill=black!25,line width=0.3pt,minimum width=2.6mm,
          minimum height=1.5mm,inner sep=0pt] at ({2.72+0.30*(\k-1)},\y) {};

\node[sm,anchor=north,text width=3.6cm,align=center] at (1.45,-4.68)
  {DAG $W{=}(V,E)$; each node tagged with tier $t$ and strategy $m$. Blocks show entries injected at that node};
\node[tx,anchor=west,font=\footnotesize\bfseries] at (4.55,0.42) {2.~Execute each node $v$};
\node[gy,anchor=west] at (7.70,0.42) {in topological order, for every node};

\node[st] (warm) at (5.35,-0.85) {warm tier};
\node[st] (dur)  at (5.35,-2.15) {durable};
\draw[amem] (warm) -- node[sm,right,pos=0.42] {miss} (dur);
\node[bx,minimum width=17mm] (inp) at (5.35,-3.45) {prompt, tools,\\query};

\node[mem] (mem)  at (7.85,-1.35) {mem$(v)$};
\node[bx] (base) at (7.85,-3.45) {base$(v)$};
\draw[amem] (warm.east) -- ++(0.30,0) |- (mem.west);
\draw[amem] (dur.east)  -- ++(0.30,0) |- (mem.west);
\draw[ar] (inp) -- (base);
\node[sm,anchor=north] at (6.82,-1.42) {top-$K$};

\node[bx] (bm) at (10.20,-2.40) {base$(v)\oplus$mem$(v)$};
\coordinate (fan) at (8.85,-3.45);
\coordinate (jm)  at (8.85,-2.40);
\draw[line width=0.45pt,draw=memS] (mem.east) -- (8.85,-1.35) -- (jm);
\draw[line width=0.45pt] (base.east) -- (fan) -- (jm);
\draw[ar] (jm) -- (bm.west);
\fill[black] (fan) circle (0.6pt);
\fill[black] (jm) circle (0.6pt);

\node[msr] (c2) at (12.55,-1.35) {count};
\node[msr] (c1) at (12.55,-3.45) {count};
\draw[amsr] (bm.east) -- ++(0.55,0) |- (c2.west);
\draw[amsr] (fan) -- (c1.west);

\node[op,draw=msrS] (mi)  at (13.80,-2.40) {$-$};
\draw[amsr] (c2.east) -- (13.80,-1.35) -- (mi.north);
\draw[amsr] (c1.east) -- (13.80,-3.45) -- (mi.south);
\node[msrhi] (inj) at (15.15,-2.40) {inject$(v)$};
\draw[amsr] (mi) -- (inj);
\node[sm,anchor=north east,text width=3.1cm,align=right] at (14.05,-4.10)
  {two non-billable counts, ${\sim}10$\,ms per node};

\node[mdl] (llm) at (9.45,-4.85) {LLM, tier $t$};
\node[mdl] (out) at (11.85,-4.85) {output$(v)$};
\draw[amdl] (bm.south) -- ++(0,-0.55) -| (llm.north);
\draw[amdl] (llm) -- (out);
\draw[da] (out.south) -- (11.85,-5.75) -- (4.30,-5.75) -- (4.30,-0.85) -- (warm.west);
\node[gy,anchor=north,align=center] at (8.20,-5.81)
  {output written back to the warm tier --- it becomes the next node's injection};

\node[tx,anchor=west,font=\footnotesize\bfseries] at (16.55,0.42) {3.~Account};

\node[acc,minimum width=25mm] (prof) at (17.75,-1.10) {cost profiler};
\draw[densely dashed,line width=0.4pt,draw=accS] (inj.east) -- (16.20,-2.40);
\draw[densely dashed,line width=0.4pt,draw=accS] (c1.east) -- (16.20,-3.45);
\draw[densely dashed,line width=0.4pt,draw=accS] (out.east) -- (16.20,-4.85);
\draw[densely dashed,line width=0.4pt,draw=accS] (16.20,-4.85) -- (16.20,-1.10);
\draw[aacc] (16.20,-1.10) -- (prof.west);

\node[acc,minimum width=25mm,align=left] (terms) at (17.75,-2.60)
 {\begin{tabular}{@{}l@{\hspace{2.5mm}}r@{}}
   $C_{\mathrm{base}}$   & billed\\
   $C_{\mathrm{inf}}$    & billed\\
   $C_{\mathrm{inject}}$ & billed
  \end{tabular}};
\node[bx,minimum width=25mm,fill=black!3,draw=black!35,text=black!45,align=left]
 (terms2) at (17.75,-3.90)
 {\begin{tabular}{@{}l@{\hspace{2.5mm}}r@{}}
   $C_{\mathrm{miss}}$  & zero here\\
   $C_{\mathrm{accum}}$ & zero here
  \end{tabular}};
\draw[ar,draw=accS] (prof) -- (terms);
\node[sm,anchor=north,text width=3.6cm,align=center] at (17.75,-4.50)
  {per-node, per-depth and per-category cost, exactly attributed};
\draw[ar,line width=0.6pt] (3.05,-2.05) -- (4.55,-2.05);
\end{tikzpicture}
\caption{End-to-end flow of a multi-agent workflow, and where memory injection
cost enters it. \textbf{1.~Compile}: a planner decomposes the user query into a directed acyclic graph (DAG)
$W=(V,E)$, and each node is tagged with a model tier $t$ and a memory strategy $m$
before execution begins. The blocks beside each node are the entries that node will
inject: a source node injects nothing, and the count grows with depth, bounded by the
retrieval window $K_{\mathrm{default}}$. \textbf{2.~Execute}: for every node in topological order, memory
is retrieved from a warm tier of capacity $K_{\mathrm{default}}$, falling back to a durable store on a
miss, and combined with the base prompt. The prompt is assembled twice --- once without
memory and once with --- and both are tokenized with the provider's own tokenizer.
Neither count is billed; together they add roughly ten milliseconds per node against
inference latencies of eighty to three thousand milliseconds. Their difference is the
exact injected token count of Eq.~(2); the first count also yields $C_{\mathrm{base}}$. The node's output is written back to the warm tier, where it
becomes part of the next node's injection --- the mechanism that makes carried context
accumulate with depth. \textbf{3.~Account}: per-node counts flow to the cost profiler
alongside the tier and strategy in use. The miss penalty and accumulation terms are zero in this harness for the reasons given in Section~\ref{sec:tca}, and are shown greyed.}
\label{fig:flow}
\end{figure*}
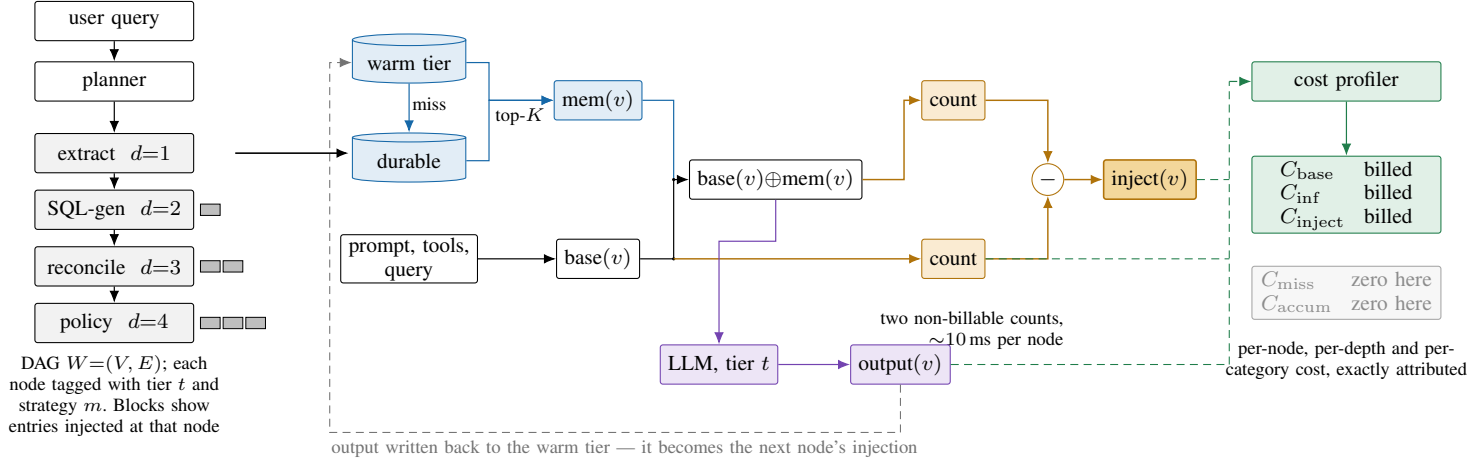

\section{Exact Injection Accounting}\label{sec:method}
Measuring $C_{\mathrm{inject}}$ requires knowing precisely how many tokens memory contributed to a prompt. Prior work that reasons about carried context estimates this quantity from word counts or character counts. Such proxies are systematically wrong by a variable margin, because subword tokenizers split text at boundaries that do not align with words, and the error depends on the content --- identifiers, numeric literals and structured output tokenize very differently from prose, and enterprise agent outputs consist largely of the former.

We measure it exactly instead, as the middle stage of Fig.~\ref{fig:flow} shows. For each node, the prompt is assembled twice: once from the base components alone, and once with the retrieved memory included. Both are tokenized, and the difference is attributed to injection:
\begin{equation}
\mathrm{inject}(v) = \mathrm{count}\big(\mathrm{base}(v) \oplus \mathrm{mem}(v)\big) - \mathrm{count}\big(\mathrm{base}(v)\big)
\end{equation}

The method has four properties that make it usable in a production deployment rather than only in a study. First, it is exact by construction: it counts the tokens the provider will bill, using the provider's own tokenizer, rather than approximating them. Second, it is free. The count operation is a non-billable API call that performs no inference and is charged as neither input nor output tokens. Third, it is cheap in latency: the two counts add roughly ten milliseconds per node against inference latencies of eighty to three thousand milliseconds, between roughly a third of one percent and an eighth of the call they instrument. Fourth, it yields $C_{\mathrm{base}}$ at the same time and at no additional cost, since $\mathrm{count}(\mathrm{base})$ is computed as the first pass; a single instrumentation point therefore produces both the injected volume and the denominator against which it should be reported.

It is worth being precise about what this measures. The method attributes; it does not predict. It runs after a node's prompt is assembled and before the call is issued, so it reports what will actually be billed rather than what a cost model expects, which distinguishes it from estimators whose accuracy depends on their priors. It says nothing counterfactual: it reports how many tokens the memory contributed, not what the node would have answered without them. Accuracy effects of removing context must be measured by execution, as in Section~\ref{sec:results}.

Integration requires no change to workflow structure: the two passes are performed by a decorator around the prompt-assembly step, which records the per-node counts alongside the tier and memory strategy in use. Because the measurement is external to the model call, it applies to any framework that assembles the prompt explicitly before dispatch and any provider exposing a tokenizer.

\section{Benchmark and Experimental Setup}\label{sec:setup}
We evaluate on a benchmark of 200 enterprise workflow tasks spanning five categories --- billing reconciliation, software asset management (SAM), identity and access management (IAM), cross-domain reconciliation and policy checking --- and five workflow depths, from two to six. The design is fully balanced: 40 tasks per category and 40 per depth, with eight tasks in every (category, depth) cell, so that per-category and per-depth comparisons are never confounded by composition (Table~\ref{tab:bench}). Distributing tasks across depths two through six lets us measure the depth dependence of injection cost directly rather than inferring it from a single workflow shape.

Workflows are composed from six specialist node classes --- extraction, present in all 200 tasks, query generation, policy checking, billing reconciliation, IAM audit and cross-domain reconciliation --- and execute against three real SQLite databases seeded with synthetic but realistic enterprise data: two hundred invoices, several hundred purchase orders and payments, and thousands of entitlement and access-log records. Token attribution is exact via the two-pass count of Section~\ref{sec:method}.

The five categories exercise different cost profiles: billing and cross-domain reconciliation are deep and dependency-heavy, SAM and IAM are query-heavy at moderate depth, and policy checking is shallow and structured. We include the last to confirm the accounting does not mistake a category needing no optimization for one that does.

\textbf{Grading.} Grading is execution-grounded and deterministic rather than delegated to a model-based rubric, because a rubric grader is itself a model with its own errors and biases. A query-generation node is graded correct only if its generated query executes successfully against the task's database, returns a non-empty error-free result, and the combined query-plus-answer text attains a recall of at least 0.6 over the task's annotated required topics. Every other node class is graded by the same required-topic recall criterion alone. End-to-end task accuracy is the verdict of the deepest node in the workflow. Where accuracy is reported at a single seed (Tables~\ref{tab:ablation} and~\ref{tab:percat}) we give no confidence interval and treat the values as directional; the three-seed headline conditions give 95\% intervals of 0.535--0.629 and 0.583--0.670, which indicates the width to expect at this benchmark size. Keyword recall is coarser than full result matching: it can credit an answer that mentions the right terms without being correct, and penalize a correct answer phrased unexpectedly. We return to this in Section~\ref{sec:limits}, and it is the reason we treat the accuracy axis of this study as a guard rather than as a result.

\textbf{Models, tiers and protocol.} All experiments use models across a small and a mid tier, with the frontier tier disabled by budget configuration so that the memory injection effect is isolated from frontier escalation. The profiler records a single frontier-tier invocation across all runs, which we report for completeness; it is one call among several thousand and affects no reported figure. Per-token prices are given in Table~\ref{tab:price}. The cost profiler is reset before each run so that routing reflects a consistent starting point, and all conditions share identical specialist prompts and retry semantics so that differences reflect the configuration under test rather than incidental execution differences. Headline conditions are run across three random seeds (42, 7 and 99); the ablation conditions are run at seed 42.

\begin{table}[t]
\caption{Benchmark composition. Eight tasks in each (category, depth) cell.}
\label{tab:bench}
\centering\footnotesize
\begin{tabular}{lcccccc}
\toprule
Category & $d{=}2$ & $d{=}3$ & $d{=}4$ & $d{=}5$ & $d{=}6$ & Total\\
\midrule
Billing & 8 & 8 & 8 & 8 & 8 & 40\\
CrossDomain & 8 & 8 & 8 & 8 & 8 & 40\\
IAM & 8 & 8 & 8 & 8 & 8 & 40\\
Policy & 8 & 8 & 8 & 8 & 8 & 40\\
SAM & 8 & 8 & 8 & 8 & 8 & 40\\
\midrule
Total & 40 & 40 & 40 & 40 & 40 & 200\\
\bottomrule
\end{tabular}
\end{table}

\begin{table}[t]
\caption{Per-token prices used throughout, in USD per million tokens (MTok)~\cite{b21}. The frontier tier was disabled by budget configuration and is not priced.}
\label{tab:price}
\centering\footnotesize
\begin{tabular}{llcc}
\toprule
Tier & Model & Input (\$/MTok) & Output (\$/MTok)\\
\midrule
Small & Claude Haiku & 1.00 & 5.00\\
Mid & Claude Sonnet & 3.00 & 15.00\\
\bottomrule
\end{tabular}
\end{table}

\section{Results}\label{sec:results}
\subsection{Injection Cost and Its Share of the Bill}
Table~\ref{tab:decomp} decomposes per-task cost into the components of Eq.~(1) for the
unoptimized baseline at the mid tier. Inference accounts for \$0.020408 per task and memory injection for \$0.003220; injection is therefore 13.6 percent of the variable cost that a compile-time optimizer can act on. Our per-condition instrumentation recorded inference and injection but not base-prompt volume, so we recover it from the cost profiler, whose observation-weighted mean is 304.4 base tokens per node at this tier over 1{,}896 observations; multiplying by the benchmark's mean of four nodes per task gives 1{,}233 tokens, or \$0.003699, and injection falls to 11.8 percent of the full billed cost. This is a derivation, not a measurement: the profiler aggregates across conditions and seeds, and the node count is the benchmark's design mean rather than a per-task observation. The 13.6 percent figure rests on direct measurement; the 11.8 percent figure is an estimate accurate to roughly a point. We report both figures because they answer different questions:
the first is the fraction of controllable cost that the memory subsystem governs, and
the second is the fraction of an invoice a practitioner would recognize.

On this benchmark the base prompt contributes slightly more billed input than injection does, 1{,}233 tokens against 1{,}073.

Injection is not the largest line in the decomposition and we do not present it as one; inference dominates at every depth we measure, and the base prompt is comparable in size. What makes injection worth attributing separately is that it is the only line that grows with the structure of the workflow rather than with the work the workflow performs.

A caution about Condition H, reported in Table~\ref{tab:ablation}. The full system (Condition H of Table~\ref{tab:ablation}) reduces the
injection line from \$0.003220 to \$0.001492, but we do not read that as a reduction in
injected volume. Condition H reassigns the majority of nodes to the small tier, and
injected tokens are billed at the input price of the tier a node lands on, so a
dollar-level change in that row conflates a change in price with a change in volume. The
two are separable only at token level and at fixed tier assignment, which is the
comparison Section~\ref{sec:results}\,C reports. We show the column for completeness and
make no optimization claim from it; Section~\ref{sec:negative} returns to why the
apparent saving belongs to tier routing rather than to memory management.

\begin{table}[t]
\caption{TCA decomposition per task for the unoptimized baseline, mid tier, seed 42.
Inference and injection are measured directly. The base-prompt row is derived rather than
measured per-condition: it is the cost profiler's observation-weighted mean of 304.4 base
tokens per node at this tier, multiplied by the mean nodes per task, and priced at
$p_{\mathrm{in}}$. See Section~\ref{sec:results}\,A for the derivation and its
limitations.}
\label{tab:decomp}
\centering\footnotesize
\begin{tabular}{lrr}
\toprule
Component & Cost per task & Share\\
\midrule
Base prompt & \$0.003699 & 13.5\%\\
Inference & \$0.020408 & 74.7\%\\
Memory injection & \$0.003220 & 11.8\%\\
Miss penalty & \$0.000000 & 0.0\%\\
Accumulation & \$0.000000 & 0.0\%\\
\midrule
Total & \$0.027327 & 100\%\\
\bottomrule
\end{tabular}
\end{table}

\subsection{Growth with Depth}
\begin{figure}[t]
\centering
\includegraphics[width=\columnwidth]{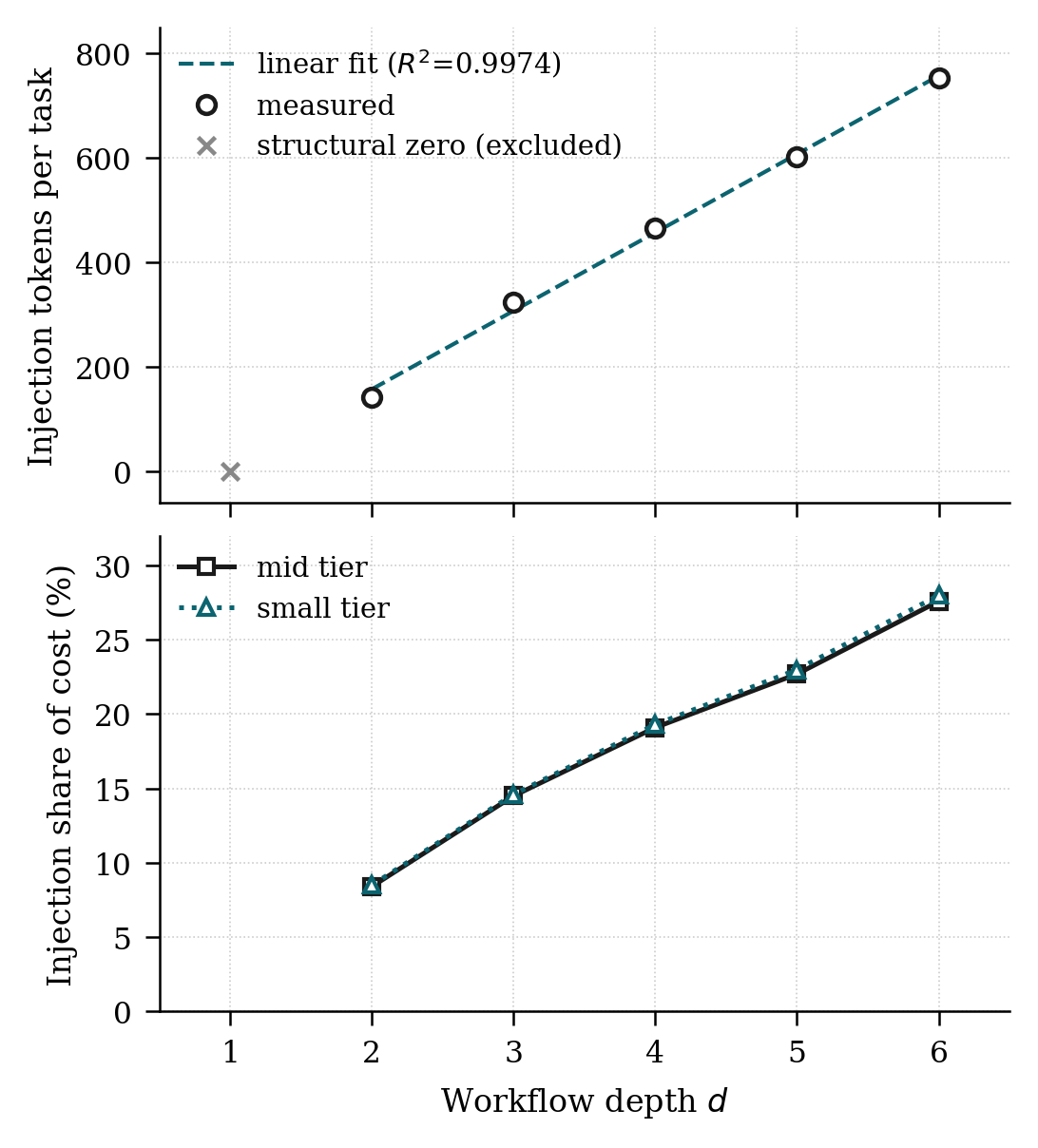}
\caption{Injection behavior across workflow depth. Top: injected tokens per task, with a linear fit over depths two through six; the depth-one point is a structural zero (a source node has no upstream output to inject) and is excluded from the regression. A quadratic fit over the same points yields a negative leading coefficient. Bottom: injection as a share of billed cost at both tiers; the curves nearly coincide because both tiers share the same input-to-output price ratio.}
\label{fig:depth}
\end{figure}

Fig.~\ref{fig:depth} reports injection behavior across workflow depth for the baseline.
The upper panel gives the absolute injected token count per task: 143 tokens at depth
two, rising through 324, 466 and 602 to 754 at depth six. A linear fit over these five
depths explains the data with $R^2 = 0.9974$ and a slope of 150 tokens per additional
level of depth. A quadratic fit over the same points reaches $R^2 = 0.9987$ with a
leading coefficient of $-4.57$: the fitted curve is concave, so the data do not merely
fail to require a quadratic term, they are inconsistent with convex growth over this
range.

The depth-one point requires comment because the benchmark contains no depth-one tasks
(Table~\ref{tab:bench}). A source node has no upstream output to inject, so its injection
is zero by the structure of the graph rather than by measurement. The instrumentation
bears this out: across 1{,}112 observations of the extraction node class at the two
active tiers, the recorded injected-token count is exactly zero in every case. We plot the point for
completeness, label it as a structural zero, and exclude it from every fit reported here;
the $R^2$ above is computed over depths two through six only. Including it changes the
linear fit to $R^2 = 0.9984$, so the conclusion does not rest on it.

The lower panel gives injection as a share of cost at each depth. The share rises
monotonically from 8.4 percent at depth two to 27.6 percent at depth six on the mid tier,
and tracks the same curve on the small tier, reaching 28.0 percent. The near-coincidence
of the two curves is itself informative: because the providers in our study price input
and output tokens in the same ratio at both tiers, the injection share is close to
tier-independent, while the absolute dollar cost of the same injected context differs by
the ratio of the two input prices. The hidden cost is therefore a structural property of
workflow depth rather than of any one tier, and its dollar impact scales with the price
of the tier it lands on.

The two panels are consistent despite one showing a rising fraction and the other a linear count: injected tokens accumulate with depth while per-node output does not, so the fraction climbs, and the retrieval window bounds per-node injection, so the count grows linearly rather than quadratically.

\subsection{Sensitivity to Retrieval Capacity}
This experiment varies one parameter at fixed tier assignment and a single seed; it shows
that the retrieval window is a live lever on injected volume, and is not a sensitivity
analysis. Table~\ref{tab:kcap}
reports the result. Reducing the warm-window capacity $K_{\mathrm{opt}}$ from its default
of 32 entries to 2, holding everything else fixed, lowers mean injected tokens per task
from 1{,}074 to 766, a reduction of 28.7 percent, and lowers cost per task from
\$0.007436 to \$0.006939, a reduction of 6.7 percent.

The token-level figure is the one that matters. Because tier assignment is unchanged
between the two rows, injected tokens are billed at the same input price in both, so the
28.7 percent reduction is attributable to the memory mechanism and to nothing else. This
is the only comparison in our experiments with that property, and it is the reason we
treat it rather than the ablation of Section~\ref{sec:negative} as the evidence that
injection cost is controllable. Cutting 28.7 percent of injected
tokens removes \$0.000308 per task, or 4.1 percent of the total; the measured reduction
is 6.7 percent. We do not claim the remaining \$0.000189 as an injection effect: a
smaller window also shortens the prompt each node conditions on, which can shorten its
output, and we did not instrument output length against capacity. The token-level
reduction is what this experiment establishes.

End-to-end accuracy falls from 0.600 to 0.570 across the same change. We do not read this
as a measured accuracy cost. A 0.030 difference sits well inside the 0.575--0.645 band
that the ablation conditions span at a single seed (Table~\ref{tab:ablation}), which we
attribute to stochastic variation rather than to any systematic effect. What two settings can establish is the sign and the order of magnitude of the tension: tightening the window reduces injected volume materially, and it eventually evicts an entry a later node needed. What they cannot establish is where the trade turns unfavorable. We therefore claim only that the lever exists and acts in the expected direction, and we specify the measurement that would locate the knee in Section~\ref{sec:limits}.

\begin{table}[t]
\caption{Retrieval capacity sensitivity. Condition B (memory only), small tier, seed 42. Token counts are per task, summed over the nodes of the workflow. This condition and tier differ from those of Fig.~\ref{fig:depth}, which reports the unoptimized baseline at the mid tier; the two are not directly comparable.}
\label{tab:kcap}
\centering\footnotesize
\begin{tabular}{cccc}
\toprule
Capacity $K_{\mathrm{opt}}$ & Inj. tokens per task & Cost per task & Accuracy\\
\midrule
32 & 1,074 & \$0.007436 & 0.600\\
2 & 766 & \$0.006939 & 0.570\\
\bottomrule
\end{tabular}
\end{table}

\section{What Did Not Work}\label{sec:negative}
We report the following results because omitting them would misrepresent the system, and because two of them bear directly on how the measurements in Section~\ref{sec:results} should be read.

\begin{figure}[t]
\centering
\includegraphics[width=\columnwidth]{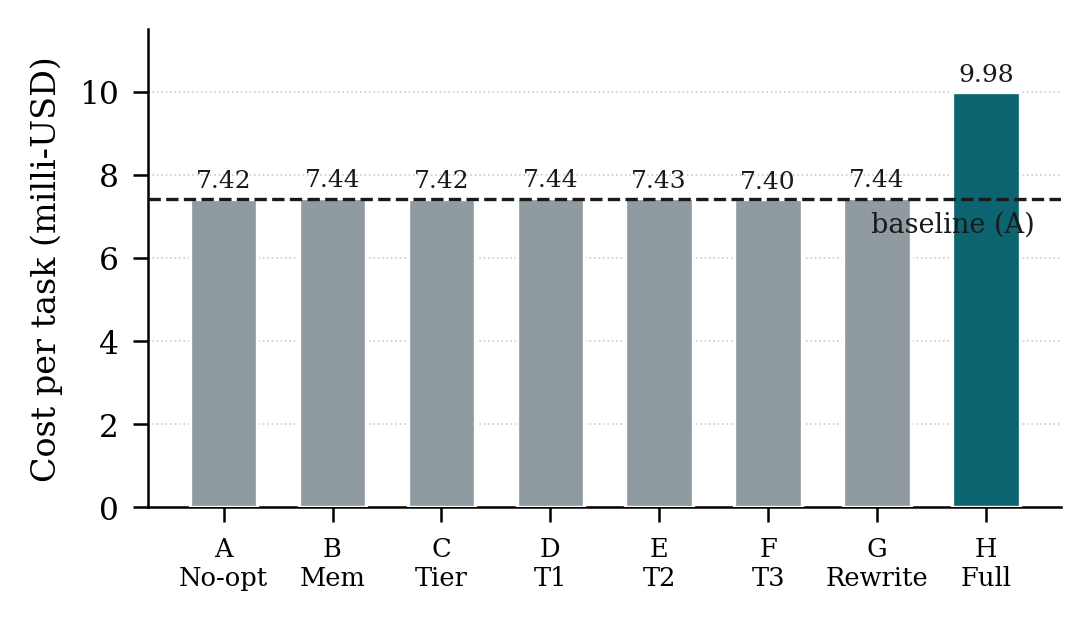}
\caption{Per-task cost by ablation condition at the small tier, 200 tasks per condition, seed 42. Conditions A through G lie within one percent of each other: at the price floor the memory and graph-rewrite transforms redistribute where context is carried without materially lowering cost. Condition H is more expensive than the baseline because its tier assigner escalates a minority of accuracy-sensitive nodes to the mid tier.}
\label{fig:ablation}
\end{figure}

\textbf{The graph-rewriting transforms are approximately cost-neutral in isolation.} Table~\ref{tab:ablation} and Fig.~\ref{fig:ablation} report per-task cost for the eight conditions at the small tier. Conditions A through G lie within one percent of each other, spanning \$0.00740 to \$0.00744 per task. Condition F, shared namespace promotion alone, is the only condition below the unoptimized baseline, and by \$0.00002; condition G, all three transforms combined, is the most expensive of the rewrite conditions rather than the cheapest. We draw the direct conclusion: at the small tier the transforms neither help nor harm cost appreciably. We do not interpret the ordering within that one-percent band. These are single-seed runs, the spread between the cheapest and most expensive of conditions A through G is \$0.00004 per task, and we have no variance estimate at this granularity; differences of that size should be read as indistinguishable rather than as a ranking. They change where context is carried without changing how much work the workflow performs.

\textbf{The full system is more expensive than the baseline at the cheapest tier.} Condition H costs \$0.00998 against the baseline's \$0.00742. This is expected rather than defective. The small tier is the price floor; there is no cheaper tier to route toward, so the tier assigner can only escalate, and it escalates the minority of accuracy-sensitive nodes to the mid tier. We record it because it makes the boundary explicit: any cost benefit from tier reassignment is available only to a deployment that starts from a more expensive tier and routes downward, which is a property of the routing prior art rather than of this article's accounting.

\textbf{Two of the five decomposition terms are zero throughout.} As stated in Section~\ref{sec:tca}, the miss penalty and accumulation terms do not accrue billable cost here, by construction rather than by omission of measurement. Warm-tier lookups do miss --- the instrumentation records a fallback on every depth-one node, and on more nodes once the window is small --- but our durable store is an in-process key-value store with no per-token billing, so a fallback costs latency and no dollars. A deployment whose durable tier is a billed remote service would populate the term. We instrument and report all five for completeness and to make explicit which two were exercised.

\textbf{Total cost is dominated by tier assignment, not by memory management.} Across our experiments the lever that moves total cost is which model each node runs on --- the finding of the routing literature~\cite{b1,b2,b6}, not of this work. We note it because memory injection is by comparison a second-order lever, and because it is why we hold tier assignment fixed in Section~\ref{sec:results}.

\begin{table}[t]
\caption{Ablation conditions, small tier, seed 42, 200 tasks each. T1 (node fusion), T2 (node reordering) and T3 (shared namespace promotion) are the three graph-rewriting transforms; condition G applies all three.}
\label{tab:ablation}
\centering\footnotesize
\begin{tabular}{clrcc}
\toprule
Cond. & Description & TCA/task & Mem \% & Acc.\\
\midrule
A & No optimization & \$0.00742 & 14.4 & 0.635\\
B & Memory only & \$0.00744 & 14.4 & 0.600\\
C & Tier only & \$0.00742 & 14.5 & 0.645\\
D & T1 fusion only & \$0.00744 & 14.5 & 0.630\\
E & T2 reorder only & \$0.00743 & 14.5 & 0.600\\
F & T3 namespace only & \$0.00740 & 14.5 & 0.575\\
G & Rewrite (all) & \$0.00744 & 14.5 & 0.610\\
H & Full system & \$0.00998 & 14.8 & 0.590\\
\bottomrule
\end{tabular}
\end{table}

\begin{table}[t]
\caption{Per-category cost and accuracy, seed 42, 40 tasks per category.}
\label{tab:percat}
\centering\footnotesize
\begin{tabular}{lrrrr}
\toprule
Category & A cost & A acc. & H cost & H acc.\\
\midrule
Billing & \$0.0234 & 0.375 & \$0.0111 & 0.600\\
CrossDomain & \$0.0239 & 0.775 & \$0.0101 & 0.825\\
IAM & \$0.0233 & 0.425 & \$0.0084 & 0.525\\
Policy & \$0.0241 & 0.750 & \$0.0096 & 0.625\\
SAM & \$0.0235 & 0.625 & \$0.0108 & 0.550\\
\bottomrule
\end{tabular}
\end{table}

\section{Limitations}\label{sec:limits}
\textbf{Prompt caching is not evaluated, and this is the most important open question.} Providers discount cached input tokens by roughly an order of magnitude~\cite{b20}, so a natural objection is that repeated injections would be served from cache and the cost we measure would largely disappear. Four structural properties work against caching injected memory here. First, a cache hit requires a byte-identical prefix up to the cache breakpoint, and injected memory consists of upstream node outputs generated within the current task instance; the first injection of such content is necessarily a write, never a read. Second, a change at the system level invalidates the message level beneath it, so in a graph whose nodes carry different specialist system prompts, memory injected at one node cannot be read from cache at another unless prompts are restructured so shared memory precedes node-specific instruction. Third, minimum cacheable prefix lengths can exceed the per-task injected context at the small tier. Fourth, content written to cache and read zero times costs more than not caching it at all, so in a fan-out graph where each node carries distinct context, naive caching raises cost. This is an argument, not a measurement: we did not instrument cache hit rates, and every figure here is for the uncached case. Measuring the hit rate on injected memory, and the residual after caching, is the first thing we would run next.

\textbf{Static DAGs.} Our benchmark comprises fixed-topology workflows; we do not evaluate ReAct-style reasoning loops~\cite{b15} or dynamically branched workflows. The accumulation mechanism is not specific to static structure: a loop at step $t$ injects the observations of steps $1$ through $t-1$, which is the accumulation of Section~\ref{sec:tca} with graph depth $d$ replaced by iteration count $t$. Because agent loops routinely run to iteration counts exceeding the depth of our deepest workflow, we would expect injection to be a larger fraction of cost in that setting. We state this as an expectation derived from the same analysis, not as a measurement.

\textbf{Tier-independence depends on a price-ratio assumption.} The near-coincidence of the two curves in Fig.~\ref{fig:depth} follows from our provider pricing input and output in the same ratio at both tiers; that is a property of one price list, not a law. Under a cheaper tier with a relatively higher input-to-output ratio the curves would separate, and under the opposite skew they would cross. Token counts are unaffected, but any share-of-cost figure must be recomputed per provider.

\textbf{Single provider, single framework.} All results use one model provider and one agent framework. The billing of input tokens is universal, and the attribution method requires only that prompts be assembled explicitly before dispatch, so we expect the accounting to transfer; whether the magnitudes do is untested.

\textbf{Grading is coarse and the accuracy axis is narrow.} End-to-end accuracy falls in a moderate band across seeds, and grading combines execution success with keyword recall over required topics at a 0.6 threshold. With 40 tasks per category, per-category differences carry wide uncertainty and should be read as directional. We use accuracy as a guard against silent degradation, not as a measured outcome, and claim no accuracy improvement anywhere in this article.

\textbf{Seed counts, and the measurement we did not make.} Headline conditions are run across three seeds; the ablation and the capacity experiment are run at a single seed. The capacity experiment reports two settings, which is enough to establish that the lever acts and in which direction, and not enough to trace the cost--accuracy curve. The specific measurement that would close this is a sweep over $K_{\mathrm{opt}} \in \{1,2,4,8,16,32,64\}$ at three seeds on the small tier under the memory-only condition, reporting injected tokens, cost and accuracy as mean and standard deviation at each setting; on our benchmark that is 21 runs of 200 tasks. We name it precisely because it is the first thing a reader should ask for, and the first thing we would run.

\section{Conclusion}
Memory injection is a real, separately attributable and exactly measurable component of multi-agent LLM workflow cost that production observability tooling does not report. We formalized it within the Total Cost of Agency decomposition, gave a two-pass token accounting method that measures it exactly at no billing cost, and characterized its behavior across workflow depth and retrieval capacity on a 200-task benchmark executed against real model APIs. We report equally that our graph transforms are approximately cost-neutral in isolation, that total cost is dominated by model tier assignment which we hold fixed, and that prompt caching --- the most plausible mitigation --- remains unmeasured and is the open question we would address first.

\bibliographystyle{IEEEtran}

\end{document}